\pdfoutput=1

\documentclass[11pt]{article}

\usepackage[preprint]{acl}
\usepackage{times}
\usepackage{latexsym}
\usepackage[T1]{fontenc}
\usepackage[utf8]{inputenc} 
\usepackage{microtype}
\usepackage{inconsolata}
\usepackage{graphicx}
\usepackage{array}         
\usepackage{multirow}      
\usepackage{rotating}      
\usepackage{booktabs}      
\usepackage{amsmath}       
\usepackage{courier}       
\usepackage{makecell}      

\usepackage{amssymb}
\usepackage[inline]{enumitem}  
\usepackage{subfigure}     
\usepackage{float}         

\newcommand{\taskscope}{\texttt{InSPEcT}}

\title{Eliciting Textual Descriptions from Representations of Continuous Prompts}

\author{Dana Ramati~~~~ Daniela Gottesman~~~~ Mor Geva \vspace{7pt} \\
  Blavatnik School of Computer Science, Tel Aviv University \vspace{7pt} \\
  {\small \texttt{ \{danaramati@mail, gottesman3@mail, morgeva@tauex\}.tau.ac.il}}
}

\begin{document}
\maketitle
\begin{abstract}
Continuous prompts, or ``soft prompts'', are a widely-adopted parameter-efficient tuning strategy for large language models, but are often less favorable due to their opaque nature.
Prior attempts to interpret continuous prompts relied on projecting individual prompt tokens onto the vocabulary space. However, this approach is problematic as performant prompts can yield arbitrary or contradictory text, and it interprets prompt tokens individually.
In this work, we propose a new approach to interpret continuous prompts that elicits textual descriptions from their representations during model inference. Using a Patchscopes variant \cite{ghandeharioun2024patchscopes} called \taskscope{} over various tasks, we show our method often yields accurate task descriptions which become more faithful as task performance increases. Moreover, an elaborated version of \taskscope{} reveals biased features in continuous prompts, whose presence correlates with biased model predictions. Providing an effective interpretability solution, \taskscope{} can be leveraged to debug unwanted properties in continuous prompts and inform developers on ways to mitigate them.
\end{abstract}

\section{Introduction}

Continuous prompts, or ``soft prompts'', are an efficient and widely-adopted solution for priming pre-trained large language models (LLMs) to solve various tasks \cite{li-liang-2021-prefix, lester-etal-2021-power}.
However, they are often less favorable compared to alternative parameter-efficient tuning methods, such as discrete prompt tuning, due to their opaque nature \cite{liu2023pre, choi-etal-2024-hard}.

How should continuous prompts be interpreted?
Prior work explored discretizing continuous prompts through projection to the model's vocabulary space \cite{khashabi-etal-2022-prompt,ju-etal-2023-continuous}. However, such approaches are problematic because it is possible to find performant continuous prompts that map to arbitrary or contradictory text \cite{khashabi-etal-2022-prompt}.
Moreover, they assume that each prompt token has an individual interpretable meaning, which does not necessarily hold.\footnote{For additional related work, please see \S\ref{sec:related_work}.}

\begin{figure}[t]
\setlength\belowcaptionskip{-8pt}
    \centering
    \includegraphics[width=\columnwidth]{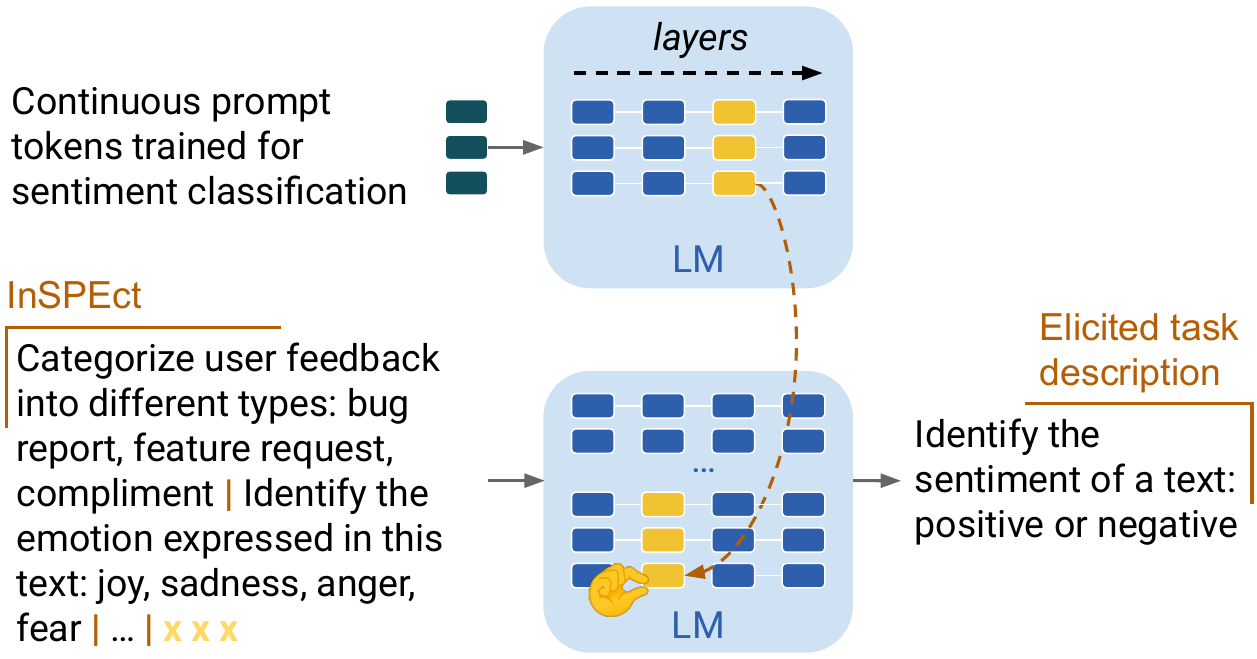}
    \caption{\taskscope{} interprets a continuous prompt by patching the prompt representations (top) into an inference pass that generates a task description (bottom).}
    \label{fig:intro}
\end{figure}

In this work, we introduce a new approach for interpreting continuous prompts that overcomes these limitations. We propose to elicit textual descriptions of the prompt from its representations, constructed by the model during inference. This is done by using the Patchscopes framework \cite{ghandeharioun2024patchscopes}; the prompt representations are extracted during inference and ``patched'' into a separate inference pass that steers the model to generate a textual description of the task (see illustration in Figure~\ref{fig:intro}).
Concretely, we define a task-description Patchscope, called \taskscope{} (Inspecting Soft Prompts by Eliciting Task descriptions), that relies on a few-shot target prompt for decoding task descriptions from patched representations. Unlike vocabulary projections that produce a discrete replacement for the prompt, \taskscope{} yields natural and easy-to-understand interpretations not bounded by the length of the prompt.

We use \taskscope{} to obtain descriptions of continuous prompts trained for 5 tasks, and find that it often yields accurate descriptions of the relevant target task (see examples in Table~\ref{tab:task_descriptions}). Generally, the higher the performance of a prompt, the more accurate the descriptions elicited from its representations.
Next, we demonstrate the utility of the elicited descriptions for debugging continuous prompts. We show that a more detailed version of \taskscope{} reveals biased features captured by prompts trained on the SNLI dataset \cite{bowman-etal-2015-large}. Moreover, when these features are present in the elicited descriptions the model exhibits biased predictions.

In summary, our work introduces a novel and practical approach for interpreting continuous prompts by eliciting natural descriptions from their representations.
We release our code at \url{https://github.com/danaramati1/InSPEcT}.

\section{Eliciting Textual Descriptions of Continuous Prompts}
\label{sec:method}

In this section, we describe our approach for interpreting continuous prompts.
Denote by $M$ a pre-trained language model with $L$ layers and hidden dimension $d$, and let ${\mathcal{P}_{cont} := \langle \mathbf{p}_1, ..., \mathbf{p}_n\rangle}$ be a continuous prompt optimized for some task $T$, where $\mathbf{p}_i \in \mathbb{R}^{d}$. We assume that $M$ is an auto-regressive transformer-based LLM \citep{vaswani} and focus on classification tasks with $C$ denoting the set of classes.

Unlike prior work  that interpreted continuous prompts by directly mapping them to discrete prompts, here we propose to elicit comprehensible descriptions of the prompts from their hidden representations constructed by $M$. We build upon the Patchscopes framework \citep{ghandeharioun2024patchscopes}, which leverages $M$'s generation capabilities to decode specific information from the model's hidden representations of some source prompt into natural language. This is done by ``patching'' the source representations into the inference pass of a different target prompt designed to encourage the extraction of that information.

\paragraph{\taskscope{}}
We treat $\mathcal{P}_{cont}$ as the source prompt that we want to interpret, and design a few-shot task-description target prompt $\mathcal{P}_{target}$:

\vspace{-8pt}
{\small 
\begin{align*}
    \texttt{``}&\texttt{desc}^{(1)}: \texttt{class}^{(1)}_1, \ldots, \texttt{class}^{(1)}_{m_1} \mid \ldots \mid \\ 
    &\texttt{desc}^{(k)}: \texttt{class}^{(k)}_{1}, \ldots, \texttt{class}^{(k)}_{m_k} \mid \texttt{x x}\ldots \texttt{x''}
\end{align*}
}%
where $\texttt{desc}^{(i)}$ is a textual description of some task $T_i \neq T$, $\texttt{class}^{(i)}_j$ is the $j$-th class label of the task $T_i$, $m_i$ is the number of classes in $T_i$, and $k$ is the number of demonstrations. The list of demonstrations is followed by a sequence of placeholder tokens (the \texttt{x}'s) for patching, of the same length as $\mathcal{P}_{cont}$. Example target prompts are shown in \S\ref{sec:target_prompts_examples}.

Denote by $\mathbf{p}_i^\ell$ the hidden  representation of $\mathbf{p}_i$ at layer $\ell$, when running $M$ on $\mathcal{P}_{cont}$. Similarly, let $\mathbf{x}_i^\ell$ be the hidden representation of the $i$-th placeholder token in the inference pass of $M$ on $\mathcal{P}_{target}$.
To elicit a textual description of $\mathcal{P}_{cont}$, we patch the representations $\mathbf{p}_1^{\ell} \ldots \mathbf{p}_n^{\ell}$ at some layer $\ell$ into the corresponding placeholder token representations $\mathbf{x}_1^{\ell^*} \ldots \mathbf{x}_n^{\ell^*}$ at some layer $\ell^*$ and let $M$ generate a sequence of tokens. If $M$ processes $P_{cont}$ as a task description, we expect it will follow the structure of $P_{target}$ and decode $P_{cont}$ into a human-readable description and set of classes for $T$. 

Eliciting textual descriptions from continuous prompts is not straightforward since there is no explicit constraint in their training that aligns them with natural language, however such interpretations could enhance their transparency and provide valuable insights into the task and features they capture.

\begin{table}[t]
\footnotesize
\setlength\tabcolsep{3.5pt}
\setlength\belowcaptionskip{-8pt}
    \centering
    \begin{tabular}{lcp{6.4cm}}
    \toprule
        & $n$ & Example elicited descriptions \\
         \midrule
         \multirow{5}{.1em}{\rotatebox{90}{SST-2}}~~ & 14  & {Identify the sentiment of a text: positive or negative} \\
          & {14} & {Categorize the tone of a text as positive, negative, or neutral}\\
          & {7} & {Identify the author's intention in this text: positive, negative or neutral} \\
         \midrule 
         \multirow{4}{.1em}{\rotatebox{90}{Subj}}~~ & {56} & {``subjective opinion or objective fact?}\\
          & {28} & {subjective, objective, or both?}\\
          & {56} & {``subjective, objective, or neutral? It is a subjective, objective, or neutral text?}\\
          \midrule 
         \multirow{4}{.1em}{\rotatebox{90}{AGNews}}~~ & {28}  & {Identify the topic of this article: technology, business, sports, world}\\
          & {56} & {Sports? Technology? Business? World?}\\
          & {28} & {World, Technology, Business, Sports, and Politics}\\
         \bottomrule
    \end{tabular}
    \caption{Accurate descriptions elicited from continuous prompts with $n$ tokens using \taskscope{} for SST-2 \citep{socher-etal-2013-recursive}, Subj \citep{pang-lee-2004-sentimental}, and AGNews \citep{NIPS2015_250cf8b5} on LLaMA2-7B-Chat.}
    \label{tab:task_descriptions}
\end{table}

\section{Experiments}
\label{sec:experiments}

We evaluate our approach and study the relationship between the interpretability and performance of continuous prompts, showing that prompts become interpretable as their performance increases.

\subsection{Experimental setting}

\paragraph{Tasks and Models} We follow \citet{khashabi-etal-2022-prompt} and base our analysis on 5 diverse classification tasks of: SST-2 \cite{socher-etal-2013-recursive} and SST-5 \cite{socher-etal-2013-recursive} for sentiment analysis, AGNews \cite{NIPS2015_250cf8b5} for news classification, Subj \cite{pang-lee-2004-sentimental} for text subjectivity, and TREC \cite{10.1145/345508.345577} for answer type classification.
As we observed low accuracy and interpretability for TREC, possibly due to the challenges of training effective prompts on this dataset \citep{min-etal-2022-noisy, khashabi-etal-2022-prompt,ju-etal-2023-continuous}, we omit it from the results.
For additional details about the tasks, see \S\ref{sec:experimental_details}.
We conduct our experiments on LLaMA2-7B-Chat \citep{touvron2023llama2openfoundation} and LLaMA3-8B-Instruct \cite{dubey2024llama3herdmodels}, each consists of 32 layers.

\paragraph{Prompt training}
For each task, we train 12 continuous prompts using standard prompt tuning \citep{lester-etal-2021-power}: 
for every prompt length $n \in \{7, 14, 28, 56\}$, we train 3 prompts with different random initializations. 
During training, we save intermediate check-points of the trainable parameters every 1K-6K examples (depending on the task and dataset size), so we can analyze the progression of task accuracy and description interpretability.
For more details, see \S\ref{sec:hyper_params}.

\paragraph{\taskscope{} demonstrations}
\label{sec:demonstrations}
In order to use \taskscope{}, we crafted a set of 8 descriptions of classifications tasks with varying numbers of classes, that are not featured in our evaluations.  
Given the sensitivity of LLMs to prompt variations \citep{min-etal-2022-rethinking, 10.1162/tacl_a_00681}, 
we interpret each continuous prompt using three target prompts with different demonstrations sampled from these task descriptions.
The set of descriptions and example target prompts are listed in  \S\ref{sec:target_prompts}.

\paragraph{Evaluation}
To assess the quality of a description $D$, we compute two metrics: 

\begin{itemize}
[itemsep=1pt, topsep=2pt,leftmargin=*]
    \item \textbf{Class Rate}: The portion of class labels present in $D$. 
    For example, in binary sentiment classification over the SST-2 dataset, if the label \texttt{positive} is present and the label \texttt{negative} is omitted in $D$, then the class rate is 0.5.

    \item \textbf{ROUGE-1}: The maximal ROUGE-1 score \citep{lin-2004-rouge} of $D$ against a set of 8-10 reference task descriptions, denoted by $\mathcal{D}_T$. Scores were computed after removing stopwords from both $D$ and the reference. To construct $\mathcal{D}_T$, we manually wrote a textual description of $T$ and then generated several paraphrased versions using ChatGPT \citep{openai_chatgpt}. For example references and generation details, see \S\ref{sec:rouge1}.

\end{itemize}
The \textit{interpretability} of a continuous prompt is measured by the average Class Rate and ROUGE-1 scores over the descriptions elicited from the three target prompts.
The \textit{performance} of the prompt is measured by the task accuracy of the model when the continuous prompt is prepended to the input example (as explained in \S\ref{sec:hyper_params}). We report evaluations on the validation sets of the SST-2 and SST-5 and the test sets of AGNews, Subj, and TREC, where validation sets are not available.

\begin{figure}[t] 
\setlength\belowcaptionskip{-12pt}
    \centering
    \resizebox{0.48\textwidth}{!}{ 
    \begin{subfigure}{}
        \centering
        \includegraphics[width=.48\textwidth]{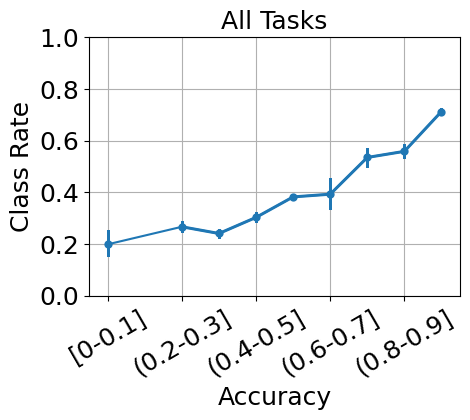}
        \label{fig:all_tasks_cr}
    \end{subfigure}
    \begin{subfigure}{}
        \centering
        \includegraphics[width=.48\textwidth]{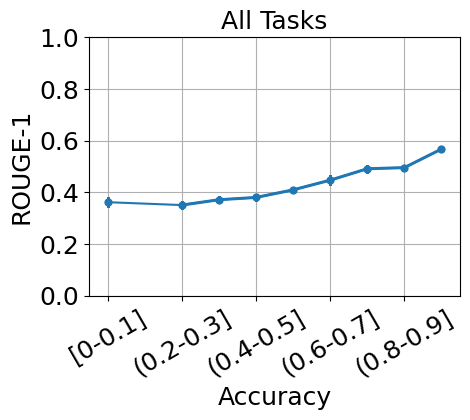}
        \label{fig:all_task_rouge}
    \end{subfigure}
    }
    \resizebox{0.48\textwidth}{!}{ 
    \begin{subfigure}{}
        \centering
        \includegraphics[width=.48\textwidth]{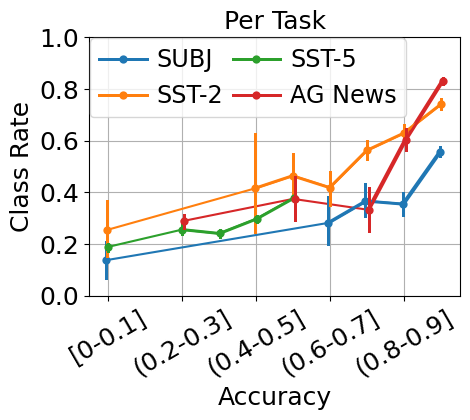}
        \label{fig:tasks_cr}
    \end{subfigure}
    \begin{subfigure}{}
        \centering
        \includegraphics[width=.48\textwidth]{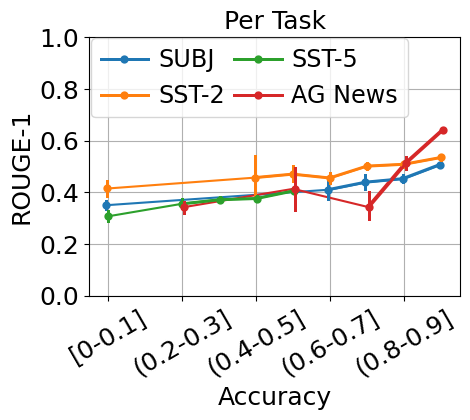}
        \label{fig:tasks_rouge}
    \end{subfigure}
    }
    \resizebox{0.48\textwidth}{!}{ 
    \begin{subfigure}{}
        \centering
        \includegraphics[width=.48\textwidth]{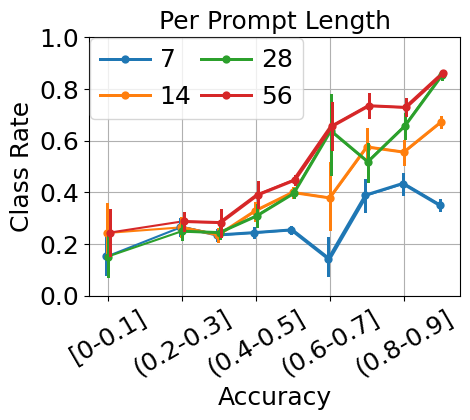}
        \label{fig:tokens_cr}
    \end{subfigure}
    \begin{subfigure}{}
        \centering
        \includegraphics[width=.48\textwidth]{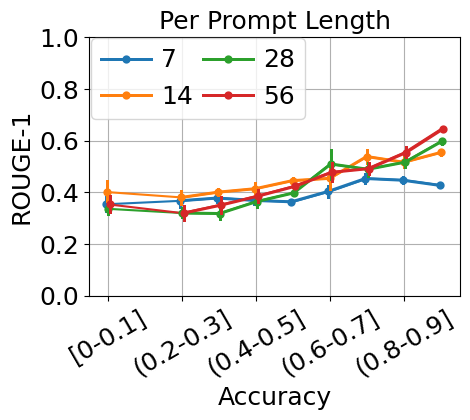}
        \label{fig:tokens_rouge}
    \end{subfigure}
    }
    \caption{Prompt interpretability as a function of task accuracy for LLaMA2. The Class Rate/ROUGE-1 scores are averaged over all the prompts within the accuracy bin. For each task and token length, the scores increase with the performance of the prompt. Results for LLaMA3 show similar trends (\S\ref{sec:additional_results}).}
    \label{fig:main}
\end{figure}

\subsection{Results}

First, we observe that our approach often elicits concise and accurate task descriptions, reaching ROUGE-1 $=$ 0.8-0.9 and covering all the task class labels (Class Rate $=$ 1.0). Examples are shown in Table~\ref{tab:task_descriptions} and in \S\ref{sec:examples}.

Next, Figure~\ref{fig:main} shows that the interpretability of a prompt increases with its task accuracy. Since elicited descriptions can be viewed as the model's interpretation of the continuous prompts, more effective continuous prompts yield more understandable and adequate descriptions. Moreover, interpretability improves as continuous prompts lengthen. We hypothesize that this trend arises because longer prompts allow the model to compress fewer task features per token \cite{elhage2022toy}.

\section{Debugging Continuous Prompts}

In this section, we demonstrate the utility of \taskscope{} for debugging continuous prompts trained over the SNLI dataset \cite{bowman-etal-2015-large}. An additional analysis for explaining the low task accuracy on SST-5 is included in \S\ref{sst5-analysis}.

\begin{figure}[t]
\setlength\belowcaptionskip{-8pt}
    \centering
    \includegraphics[width=0.9\columnwidth]{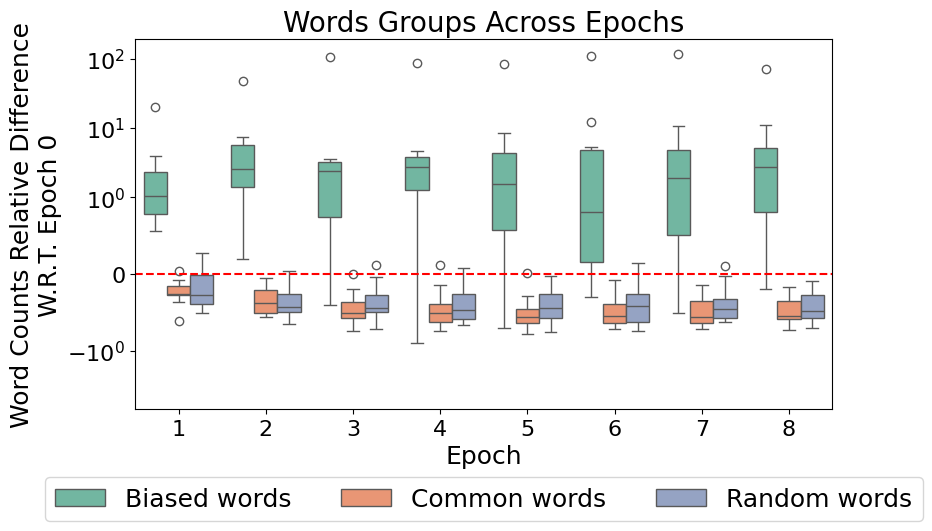}
    \caption{Differences in counts of each word group in generated outputs during training with respect to randomly-initialized prompts (epoch~0). 
    The distributions are aggregated over 10 continuous prompts trained on SNLI \cite{bowman-etal-2015-large}.}
    \label{fig:snli}
\end{figure}

\paragraph{Eliciting spurious correlations using \taskscope{}}
The SNLI dataset is known to have multiple biases \citep{gururangan-etal-2018-annotation, mersinias-valvis-2022-mitigating} which allow models to learn shortcuts, such as the correlation of negation and vagueness with certain classes. 
We use SNLI to train 10 different 14-token continuous prompts, check-pointed over 8 epochs, which vary in random initialization. \taskscope{} is applied to each check-point using a target prompt that elicits the learned features: 

\vspace{-12pt}
{\small
\begin{align*}
    & \texttt{``Respond with a short sentence. What} \\
    & \texttt{features are used for classifying each} \\
    & \texttt{label in the following task: x x}\ldots\texttt{x''}
\end{align*}
}%
Next, we count the appearances of distinct word groups in the generated outputs: (a) \textit{biased words}: words with top-5 highest spurious correlations per class, as reported in \citet{Wu2022GeneratingDT} Table~12, (b) \textit{common words} (baseline): top-10 most frequent words across all generated outputs, omitting stopwords, digits and words in the target prompt, and (c) \textit{random words} (baseline): 10 words randomly sampled from all generated outputs.
For each generated output and group, we measure the word count difference with respect to the output of a randomly initialized prompt (epoch 0).

Figure~\ref{fig:snli} shows that biased words indeed emerge in the generated outputs from the beginning of training, indicating the presence of these features in SNLI continuous prompts. This contrasts the appearance of the baseline word groups, which slightly decrease.

\begin{figure}[t]
\setlength\belowcaptionskip{-8pt}
    \centering
    \includegraphics[scale=0.3]{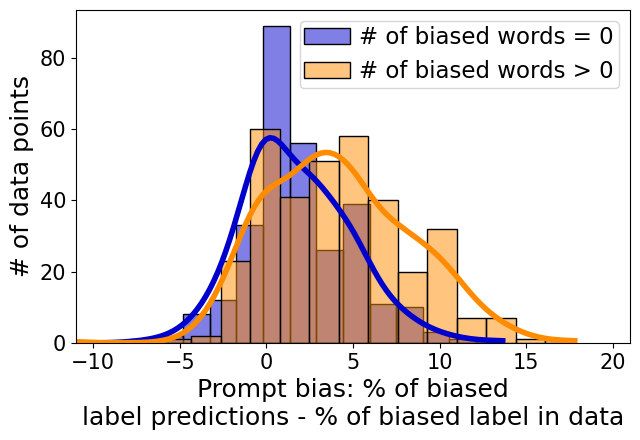}
    \caption{Histograms of the counts of generated biased words across different prompt bias levels. Outputs with biased words $(>0)$ show positive predictive bias, while those without $(=0)$ are unbiased on average. The x-axis is cut to $[-10, 20]$ for brevity, omitting outliers.}
    \label{fig:bias_dist}
\end{figure}

\paragraph{Elicited biases correlate with biased predictions}
If continuous prompts indeed capture the biases elicited from \taskscope{}, then we expect them to encourage biased model predictions.
To test this, we take each continuous prompt check-point and biased word pair, and quantify the model's predictive bias towards the bias-correlated class.
Bias is measured by calculating the percentage difference between predicted and actual cases of a bias-correlated class among dataset examples containing the biased word, with larger differences indicating higher predictive bias. For example, considering the biased word \texttt{outside} and bias-correlated class \texttt{entailment} --- if $65\%$ of the relevant examples are true entailment cases and $70\%$ are predicted as such, the bias measure is $5\%$.

Figure~\ref{fig:bias_dist} shows that predictive bias is generally positive for outputs containing biased words, and centered around 0 for outputs lacking them. A sign test comparing these distributions indicates significantly higher predictive bias when a biased word is present (p-value $2.96e^{-11}$).

\section{Conclusion}
We tackle one of the major hurdles of continuous prompts --- their lack of transparency. 
We show that accurate task descriptions can be elicited with \taskscope{} from the model's internal representations, and task performance improves as the model's own interpretation of the prompts becomes more faithful.
Additionally, \taskscope{} can identify biased features in continuous prompts from the presence of prominent words in the generated outputs. Overall, our work provides an effective interpretability solution that can be leveraged to debug unwanted properties in continuous prompts.

\section*{Limitations}

Following previous work on interpretability of continuous prompts \cite{khashabi-etal-2022-prompt, ju-etal-2023-continuous}, our experiments focus on classification tasks where evaluation is easier compared to open-ended generation tasks.
Extending our analyses to other tasks is an interesting direction for future work.

We were often able to elicit meaningful and understandable task descriptions, though there were some the cases where the descriptions were unclear and did not yield informative content, especially early in training. Since \taskscope{} can be viewed as the model’s interpretation of the continuous prompts, identifying the precise conditions for its success may align with optimizing training configurations that enable the model to learn more effectively.

Our work explores the correlation between prompt interpretability and task performance by finding a meaningful one-way mapping from continuous prompts to discrete forms. Conducting a causal analysis --- where elicited descriptions are modified, mapped back to continuous prompts, and evaluated for changes in task performance --- could offer deeper insights into how models use and form predictions based on the information encoded within continuous prompts. 

Prior work focused on discretizing continuous prompts such that the discrete prompts can be used as replacements that yield equivalent task performance and class label prediction distributions. Notably, our method does not produce such discrete replacements, but rather elicits information in a textual and easy-to-understand format to better understand the information encoded in the continuous prompt and its potential for debugging.
While we found the elicited descriptions to be generally informative and accurate, they do not necessarily guide the model to produce explicit class labels like their corresponding continuous prompts.

\section*{Acknowledgments}
We thank Ori Yoran, Asma Ghandeharioun, and Katja Filippova for constructive feedback. This research was supported in part by The Israel Science Foundation grant 1083/24, Len Blavatnik and the Blavatnik Family foundation.

\bibliography{custom}

\appendix

\section{Additional Experimental Details}
\label{sec:experimental_details}

\subsection{Downstream Tasks}
\begin{table}[H]
\footnotesize
\centering
    \begin{tabular}{llc}
        \toprule
        Dataset & Task &  $|C|$ \\
        \midrule
        AGNews & News topic classification & 4 \\ 
        SST-2   & Sentiment analysis (movie) & 2 \\ 
        SST-5   & Sentiment analysis (movie) & 5 \\ 
        Subj  &  Subjectivity classification & 2 \\
        TREC    &  Answer type classification & 6 \\  
        \bottomrule 
    \end{tabular}
    \caption{    
    The set of downstream tasks used in the experiments, where $|C|$ represents the number of classes for each task.
    }
    \label{tab:tasks}
\end{table}

\subsection{Training Details}
\label{sec:hyper_params}

Given a training set $\mathcal{D} = \{(x_i, y_i)\}_{i=1}^{|\mathcal{D}|}$, where $x_i$ is an input text for classification and $y_i$ is a gold class label, $\mathcal{P}_{cont}$ is learned by minimizing the cross-entropy loss between $y_i$ and the model's predicted label for the input ${\texttt{``}\mathbf{p}_1 \ldots \mathbf{p}_n \ \texttt{Text:} [x_i]\texttt{, Label:''}}$ over $\mathcal{D}$.

\begin{table}[H]
\footnotesize
\centering
    \begin{tabular}{lccc}
        \toprule
        Dataset & Learning Rate & Epochs & \(|T|\) \\
        \midrule
        AGNews & {$8e^{-3}$} & 8  & 50,000 \\ 
        SST-2  & {$8e^{-4}$} & 8  & 50,000 \\ 
        SST-5  & {$6e^{-3}$} & 12 & 8,500  \\ 
        Subj   & {$8e^{-3}$} & 8  & 8,000  \\
        TREC   & {$8e^{-4}$} & 20 & 5,400  \\  
        \bottomrule 
    \end{tabular}
    \caption{    
    Hyper-parameters used to train prompts on both LLaMA2 7B chat and LLaMA3 8B Instruct models. \(|T|\) represents the size of the training set used.
    }
    \label{tab:params}
\end{table}

\subsection{Resources}
\label{sec:resources}

All our experiments were conducted using a single A100 80GB or H100 80GB GPU.

\subsection{Software Packages}
\label{sec:packages}

We used the PyTorch Python package \citep{NEURIPS2019_9015} for training the continuous prompts and conducting the experiments. For calculating the scores, we used the rouge-score Python package \citep{google-research-rouge} for ROUGE-1, and the NLTK Python package~\citep{bird-loper-2004-nltk} for removing English stopwords, both with default parameters.

\section{Example Interpretations of Continuous Prompts}
\label{sec:examples}

Examples of discrete prompts elicited using \taskscope{} on LLaMA2-7B-Chat and LLaMA3-8B-Instruct are presented in Table~\ref{tab:discrete_examples_llama2} and Table~\ref{tab:discrete_examples_llama3}, respectively.

\begin{table*}[t]
    \centering
    \footnotesize
    \setlength\tabcolsep{5pt} 
    \setlength\belowcaptionskip{-8pt}
    \begin{tabular}{lp{15cm}} 
    \toprule
        \multirow{12}{.1em}{\rotatebox{90}{AG News}}~~ & number, as well as the latest news and updates from the world of technology \\ 
         & with, Digital Marketing, Business, and Technology topics \\ 
         & Identify the main topic of this text: technology, entertainment, politics, sports \\ 
         & Identify the main theme of the text: technology, business, politics \\ 
         & Club, or Identify the topic of this text: entertainment, politics, sports, or technology \\ 
         & ? technology \&? business  \&? entertainment  \&? sports  \&?  World  \&? news  \&? lifestyle  \& \\ 
         & world, Technology, Business, and Sports \\  
         & -world, the following categories: Sports, Business, Technology, Entertainment, and Science \\ 
         & -- World-- Technology-- Business-- Sports-- World \\ 
         & Sports? Technology? Business? World news? We will be happy to help you with any question you have! \\ 
         & Technology World Business Sports \\
         & is, World, Sports, Business, Technology \\
         \midrule
        \multirow{12}{.1em}{\rotatebox{90}{SST-2}}~~ & xtake a look at the text and identify the tone: positive, negative, or neutral \\ 
         & give feedback on a product: positive, negative, or neutral \\ 
         & Identify the sentiment of a text: positive, negative, or neutral \\ 
         & Categorize the tone of a text as positive, negative, or neutral \\ 
         & ance as a positive or negative response? \\ 
         & and negative sentiment? \\ 
         & Please note that the text is a positive or negative? \\ 
         & U ( positive) and U (negative) are used to indicate the emotions expressed in the text \\ 
         & ? a positive or negative review?000000000000000000000000 \\ 
         & ES of negativity, but positivity? \\ 
         & Identify the tone of a piece of writing: positive, negative, neutral \\ 
         & rices and negative feelings, but also positive feelings, such as joy, happiness, and contentment \\ 
         \midrule
         \multirow{18}{.1em}{\rotatebox{90}{SST-5}}~~ & leaving feedback on a product or service: good, bad, or neutral \\ 
         & yeah (yes) (great job) (excellent) (good work) (well done) (superb) (amazing) \\ 
         & (A) great (B) good (C) okay (D) poor \\ 
         & yeah (100\%), great (80\%), okay (60\%), meh (40\%), bad (20\%  \\ 
         & anarchy Is this a good or bad thing? \\ 
         & yevaluate the quality of a piece of writing: good, neutral, or bad \\ 
         & by: good, neutral or bad \\ 
         & -ilk to which it is assigned: good, bad or neutral \\ 
         & : This is a good or bad thing: Neutral \\ 
         & by which I would classify it: good or bad \\
         & bad? Very bad? Worse than bad? Terrible? Horrible? Abysmal? \\ 
         & bad, my dear, this is a great answer \\ 
         & nough, great, good, bad, or ugly?t is a genre of literature that explores the impact of science and technology on society \\ 
         & testing is okay, but not great, but not terrible, but not good \\ 
         & -based on their answers: good, neutral, or bad \\ 
         & say goodness, the text is neutral \\ 
         & bad and terrible) and 50/50 chance of being a good or bad review \\ 
         & not enough this topic not enough to be considered as a good or bad reviews? \\
         \midrule
        \multirow{15}{.1em}{\rotatebox{90}{SUBJ}}~~ & (Learning Objective 1) \\ 
         & matter of fact, opinion or perspective \\ 
         & coverage of a news article or event: objective, subjective, persuasive \\ 
         & matter of factuality or subjectivity \\ 
         & The above are examples of subjective and objective criteria for evaluating the quality of a text or author \\ 
         & `s the subjective and objective' \\ 
         & subjective opinion of the matter \\ 
         & subject to the subjective opinion of the observer \\ 
         & The term ``subjective'' refers to something that is based on personal opinions or preferences, rather than objective facts \\ 
         & The answer to this question depends on how you define ``subjective'' and ``objective'' are two different things \\ 
         & subjective, objective, or both?  \\  
         & in this passage, but the subjective and objective, \\ 
         & the objective of this exercise is to assess the subjective value of the answer \\ 
         & in a subjective, subjective, objective, or objective manner \\ & ``objective''``subjective''``opinion''``fact'' \\       
    \bottomrule
    \end{tabular}
    \caption{Examples of accurate task descriptions elicited using \taskscope{} on LLaMA2-7B-Chat.}
    \label{tab:discrete_examples_llama2}
\end{table*}

\begin{table*}[t]
    \centering
    \footnotesize
    \setlength\tabcolsep{5pt} 
    \setlength\belowcaptionskip{-8pt}
    \begin{tabular}{lp{15cm}} 
    \toprule
        \multirow{11}{.1em}{\rotatebox{90}{AG News}}~~ & World Technology Business \\ 
         & nikaite Technology; or a Business; or Entertainment; or Sports; or Sports; or Technology; or Technology; \\ 
         & 279; in Business; Sports; Entertainment; Technology;; \\ 
         & Technology; Business; And World \\ 
         & -including the World of Business or Leisure or Sports or Technology or News or Culture or Healthassistant; \\ 
         & Technology; to classify this passage from: business \\ 
         & Worldwide in a World of Business or Technology or Entertainment or Health or Fashion or Sports or World \\  
         & and answer: What is the main topic of this article?assistant \\ 
         & Identify the type of the website: Technology, Entertainment, Sports, Business \\ 
         & /oriented to the world of sports, the text is a sports news article \\ 
         & ultiimateley, classify this text into a genre: business, technology, entertainmentassistant \\
         \midrule
        \multirow{13}{.1em}{\rotatebox{90}{SST-2}}~~ & The text is a positive or negative: \\ 
         & Identify the positive and negative statements in a text \\ 
         & Identify the positive/negative emotions in a text: positive, negative \\ 
         & Identify the positive or negative sentiment of a text \\ 
         & Identify the positive and negative aspects of a text: The positive aspects of a text: The negative aspects of a text: \\ 
         & Determine the sentiment of a text: positive, negative, or neutral \\ 
         & lettered a positive or negative \\ 
         & The text is a negative review of a movie, which is a negative review \\ 
         & From a book: Identify the author's tone: positive, negative, formal, informal, sarcastic, or philosophical \\ 
         & ://positive-negative-negative \\ 
         & Is this a positive or negative review: positive, negative \\ 
         & Is this sentence a positive or negative statement \\ 
         & Categorize this text into a category: positive, negative, neutral \\
         \midrule
         \multirow{13}{.1em}{\rotatebox{90}{SST-5}}~~ & badgered = 2;terrible = 2;good = 2;neutral = 2;bad = 2;terrible = 2;good = 2;neutral = 2;bad = 2;terrible = 2;good = 2; \\ 
         & neutral of the good or bad of the game \\ 
         & Identify the author of this text:terrible, good, neutral \\ 
         & Answer: The text: a neutral good: a good'totalitarian a: a bad: aterrible:terrible:terrible:terrible \\ 
         & onenasty of the text: neutral, good or bad \\ 
         & ://good or bad text \\ 
         & terrible, awful, bad, good, excellent, great, wonderful, lovely, beautiful, lovely, lovely \\ 
         & :bad news, neutral, good news, neutral, bad news, good news, bad news \\ 
         & :good or bad \\
         & Is the information in this sentence good or bad? \\ 
         & Is it a good news, bad news, or neutral news \\ 
         & idiagnosis, a good or bad, and neutral \\ 
         & I cannot be used, a good, neutral, or bad \\ 
         \midrule
        \multirow{13}{.1em}{\rotatebox{90}{SUBJ}}~~ & Identify the tone of this text: formal, informal, formal and objective, formal and subjective \\ 
         & Objective Subjective Subjective \\ 
         & objective and subjective language: objective language is used to describe the facts, while subjective language is used~to- express the author's opinion or feeling \\ 
         & Objective of the learning objectives of the Subjective Subjective \\ 
         & Objective: The text of the subjective \\ 
         & The text is a subjective and/or objective and/or subjective/objective \\ 
         & Objective: To identify the emotion expressed in the text \\ 
         & Identify the subject of a text: objective, subjective  \\  
         & Please note that the classification is subjective and may not be objective \\ 
         & ://mannerisms of a text: Identify the tone of a text: objective, subjective, formal, informal, sarcastic \\  
         & ://determine the tone of the text: objective, objective, objective \\ 
         & Subjective: The text is subjective as it is a subjective text \\       
    \bottomrule
    \end{tabular}
    \caption{Examples of accurate task descriptions elicited using \taskscope{} on LLaMA3-8B-Instruct.}
    \label{tab:discrete_examples_llama3}
\end{table*}

\section{Additional Results}
\label{sec:additional_results}

\begin{figure}[H]
    \centering
    \resizebox{0.48\textwidth}{!}{ 
    \begin{subfigure}{}
        \centering
        \includegraphics[width=.48\textwidth]{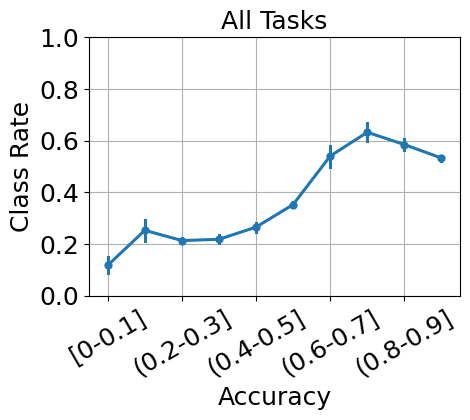}
        \label{fig:all_tasks_cr_3}
    \end{subfigure}
    \begin{subfigure}{}
        \centering
        \includegraphics[width=.48\textwidth]{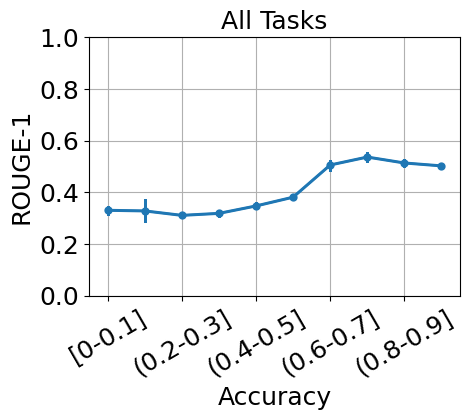}
        \label{fig:all_task_rouge_3}
    \end{subfigure}
    }
    \resizebox{0.48\textwidth}{!}{ 
    \begin{subfigure}{}
        \centering
        \includegraphics[width=.48\textwidth]{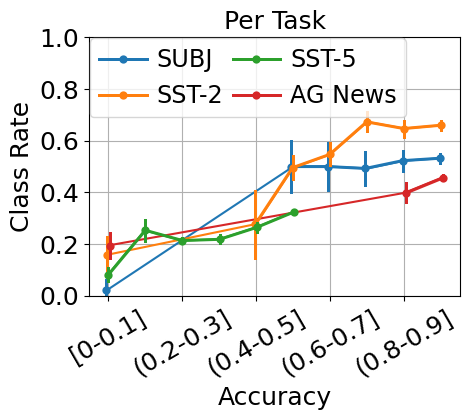}
        \label{fig:tasks_cr_3}
    \end{subfigure}
    \begin{subfigure}{}
        \centering
        \includegraphics[width=.48\textwidth]{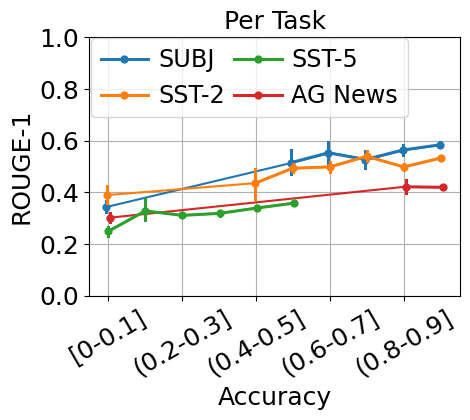}
        \label{fig:tasks_rouge_3}
    \end{subfigure}
    }
    \resizebox{0.48\textwidth}{!}{ 
    \begin{subfigure}{}
        \centering
        \includegraphics[width=.48\textwidth]{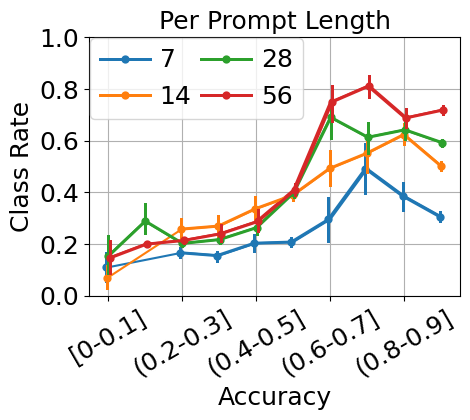}
        \label{fig:tokens_cr_3}
    \end{subfigure}
    \begin{subfigure}{}
        \centering
        \includegraphics[width=.48\textwidth]{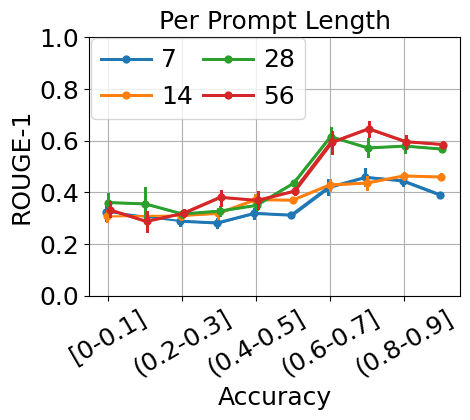}
        \label{fig:tokens_rouge_3}
    \end{subfigure}
    }
    \caption{Prompt interpretability as a function of task accuracy for LLaMA3. The Class Rate/ROUGE-1 scores are averaged over all the prompts within the accuracy bin.}
    \label{fig:additional_results}
\end{figure}

The results for LLaMA3 are presented in Figure~\ref{fig:additional_results}. We observe similar trends to those of LLaMA2. First, we observe that the interpretability of a prompt improves as its task accuracy increases. However, there is a small drop in interpretability within the 0.8 to 1 accuracy range, likely due to the trends observed across all tasks when using 7 tokens, affecting both class rate and ROUGE-1 scores. Additionally, interpretability improves as continuous prompts lengthen, as obsereved in LLaMA2.

\section{Target prompts}
Examples of task descriptions and target prompts are presented in this section. Discussions regarding their use and generation are in \S{\ref{sec:method}} and \S\ref{sec:demonstrations}, respectively.

\label{sec:target_prompts}
\subsection{Crafted Classification Tasks Descriptions}

The following task descriptions were used for sampling and constructing target prompts:

\begin{itemize}[noitemsep]
    \item Identify the emotion expressed in this text: joy, sadness, anger, fear
    \item Is the information in this sentence correct?: True, False
    \item Classify this passage from a book or movie into its genre: science fiction, romance, thriller
    \item Determine who is the author of a given text: Shakespeare or Marlowe
    \item Identify which season is described in this text: summer, winter, autumn or spring
    \item Categorize customer feedback into different types: bug report, feature request, compliment
    \item Identify the type of this email: spam or not spam
    \item Identify the political leaning of a text or author: left or right
\end{itemize}

\subsection{Example Target Prompts}
\label{sec:target_prompts_examples}
The following target prompts were used by \taskscope{} to elicit task descriptions:
\begin{itemize}[noitemsep]
    \item Categorize customer feedback into different types: bug report, feature request, compliment | Identify the emotion expressed in this text: joy, sadness, anger, fear | Is the information in this sentence correct?: True, False | x x x x x x x
    \item Determine who is the author of a given text: Shakespeare or Marlowe | Categorize customer feedback into different types: bug report, feature request, compliment | Identify the political leaning of a text or author: left or right | x x x x x x x x x x x x x x
    \item Identify the emotion expressed in this text: joy, sadness, anger, fear | Classify this passage from a book or movie into its genre: science fiction, romance, thriller | Identify which season is decsribed in this text: summer, winter, autumn or spring | x x x x x x x x x x x x x x x x x x x x x x x x x x x x
\end{itemize}

\section{ROUGE-1 Calculation Details}
\label{sec:rouge1}
Further details regarding the computation of the ROUGE-1 scores are discussed below.

\subsection{Stopwords Removal}
 To prevent computing misleadingly high ROUGE-1 scores for discrete prompts that closely resemble the format of reference descriptions, but fail to accurately capture the target task, we removed stopwords from both the elicted \taskscope{} descriptions and the reference descriptions in Table~\ref{tab:reference_descriptions}. This was achieved using the NLTK Python package \citep{bird-loper-2004-nltk}.

\subsection{References Descriptions Creation}
To compute the final ROUGE-1 score for each description \(D\) elicted by \taskscope{}, we used ChatGPT to generate 8-10 reference descriptions per task. The input format we used to prompt ChatGPT was: ``\texttt{Could you rephrase the following sentence and provide a few options: <SENTENCE>}'', where \texttt{<SENTENCE>} represents a brief description of the target task.
Examples of reference descriptions generated are presented in Table~\ref{tab:reference_descriptions}.

\begin{table*}[ht]
\centering
\footnotesize
\setlength\tabcolsep{5pt} 
\setlength\belowcaptionskip{-8pt}
\begin{tabular}{lp{15cm}} 
\toprule
    & Example reference descriptions \\
    \midrule
    \multirow{5}{.1em}{\rotatebox{90}{AGNews}}~~ & {Which topic is this article about? World, Sports, Business, Technology}\\
    & {What is the main topic discussed in this article: World, Sports, Business, Technology}\\
    & {What is the most fitting summary for this article? World, Sports, Business, Technology}\\
    & {Among World, Sports, Business, and Technology, which best captures the topic of this article}\\
    & {To which category does this news article's topic belong: World, Sports, Business, Technology}\\
    \midrule
    \multirow{5}{.1em}{\rotatebox{90}{SST-2}}~~ & {Is the sentiment of this sentence positive or negative?} \\
    & {Would you classify this sentence as having a positive or negative sentiment?}\\
    & {Can you identify whether the sentiment of this sentence is positive or negative?} \\
    & {Would you consider the sentiment of this sentence to be positive or negative?} \\
    & {What is the tone of this sentence: positive or negative?} \\
    \midrule 
    \multirow{5}{.1em}{\rotatebox{90}{SST-5}}~~ & {Is the sentiment of this sentence terrible, bad, neutral, good or great?}\\
    & {Do you think this sentence has a terrible, bad, neutral, good or great tone?}\\
    & {How would you rate the sentiment of this sentence: terrible, bad, neutral, good or great?}\\
    & {How would the sentiment of this sentence be described? terrible, bad, neutral, good, great.}\\
    & {Would you classify this sentence as having a terrible, bad, neutral, good or great sentiment?}\\
    \midrule
    \multirow{5}{.1em}{\rotatebox{90}{Subj}}~~ & {Is the subjectivity of this text objective or subjective?}\\
    & {In terms of subjectivity, is this sentence objective or subjective?}\\
    & {Classify the sentence based on its expression: objective, subjective}\\
    & {Is this sentence objective or subjective in nature?}\\
    & {Determine if this sentence presents facts or opinions: objective, subjective}\\
    \midrule
    \multirow{5}{.1em}{\rotatebox{90}{TREC}}~~ & {Is the question asking about an entity, a description, an abbreviation, an expression, a human, a location, or a number?}\\
    & {What type of thing is the question asking about? Description, entity, abbreviation, expression, human, location, number}\\
    & {What type is the answer to this question: entity, description, abbreviation, expression, human, location, or number?}\\
    & {Choose the category that best fits the answer: Description, Entity, Abbreviation, Expression, Human, Location, Number}\\
    & {Does the question pertain to an entity, a description, an abbreviation, an expression, a human, a location, or a number?}\\
\bottomrule
\end{tabular}
\caption{Example of reference descriptions used to calculate ROUGE-1 scores.}
\label{tab:reference_descriptions}
\end{table*}

\section{Debugging Low Task Performance}
\label{sst5-analysis}
\begin{figure}[t]
    \centering
    \includegraphics[width=0.85\columnwidth]{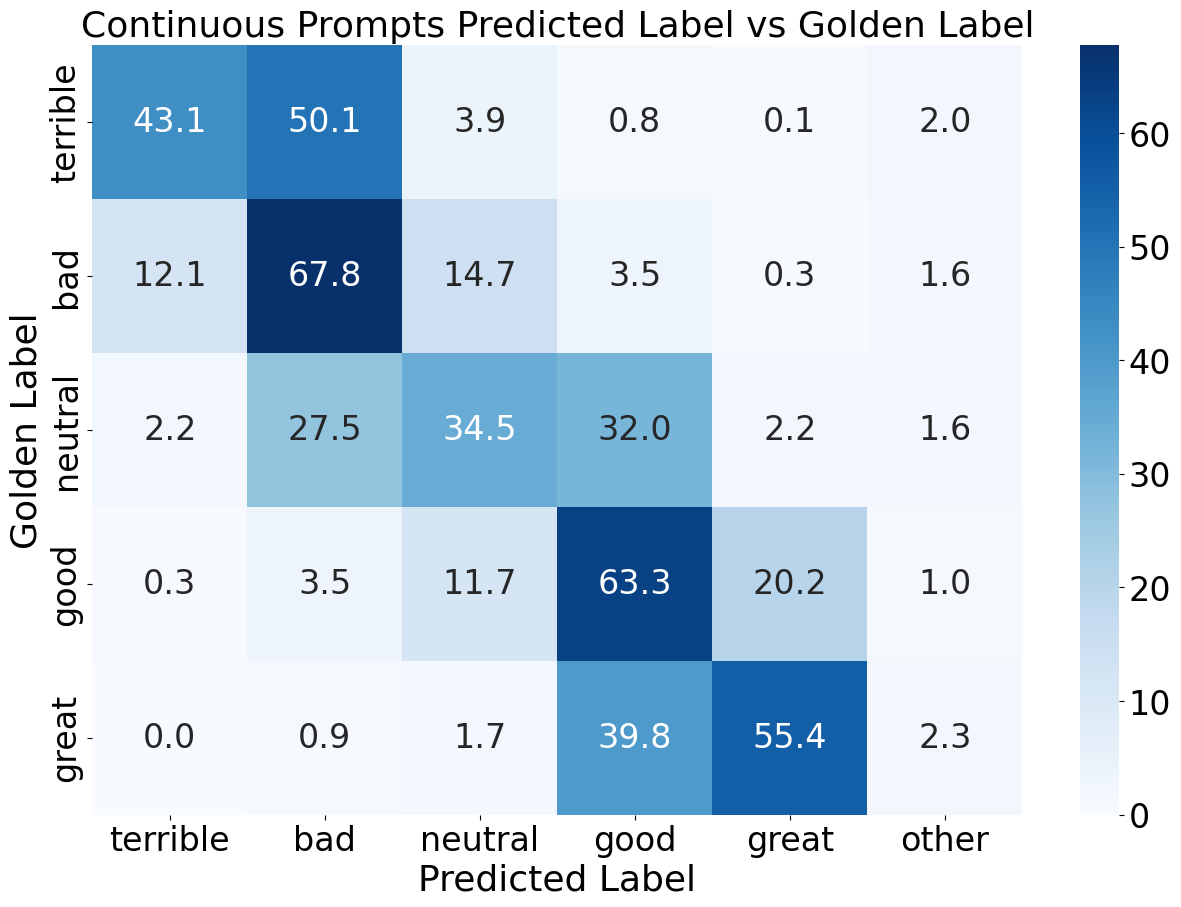}
    \caption{Confusion matrix comparing predictions generated by continuous prompts which captured only three classes, and the true labels.}
    \label{fig:sst5}
\end{figure}

In the SST-5 dataset, the trained continuous prompts achieved $50\% - 60\%$ task accuracy. Tables~\ref{tab:discrete_examples_llama2} and~\ref{tab:discrete_examples_llama3} contain examples of elicited \taskscope{} descriptions, which often list only a subset of class labels: ``\texttt{good}'', ``\texttt{bad}'', ``\texttt{neutral}''. A possible explanation for the poor performance is that the continuous prompt steers the model to produce only a partial set of classes. Figure~\ref{fig:sst5} presents a confusion matrix, with values representing dataset example counts, between the predictions generated by continuous prompts where the elicited descriptions captured only three classes, and the true labels. These prompts struggled to capture the nuanced differences between ``\texttt{good}'' and ``\texttt{great}'', shown by the similar prediction rates $39.8\%$ and $55.4\%$ for examples from the ``\texttt{great}'' class. Similar confusion is demonstrated for examples from the ``\texttt{terrible}'' class, where prediction rates are $43.1\%$ and $50.1\%$ for ``\texttt{terrible}'' and ``\texttt{bad}'', respectively. The omission of the difficult classes in the \taskscope{} descriptions could indicate that the continuous prompts may not recognize the full spectrum of sentiment represented in SST-5.

\section{Related Work}
\label{sec:related_work}
\paragraph{Interpreting continuous prompts}
Interpreting continuous prompts has been attempted by projecting individual prompt tokens to the vocabulary space \citep{khashabi-etal-2022-prompt, webson-pavlick-2022-prompt} or by optimizing an external objective to map them to their discrete forms \citep{ju-etal-2023-continuous}. However, these mappings operate on each token individually, often contain several noisy tokens that are difficult to understand \citep{ju-etal-2023-continuous}, and may yield discrete interpretations that are irrelevant or contradictory \citep{khashabi-etal-2022-prompt}.

\paragraph{Embedding inversion}
Previous research has investigated reconstructing text from dense representations by learning a function that inverts the text encoder \citep{morris-etal-2023-text}. Other approaches identify which content activates certain model components in order to decipher the information encoded in new inputs \citep{huang2024inversionview}. These methods involve extensive analysis and rely on external optimizations. In contrast, our approach simply leverages the model's intrinsic generation capabilities to form understandable descriptions of continuous prompt embeddings.

\paragraph{Bias in continuous prompts}
Models may rely on spurious correlations between classes and specific words \citep{Wu2022GeneratingDT}, and superficial clues \citep{kavumba-etal-2022-prompt}, like high lexicographical overlap between the premise and hypothesis in natural language inference, to perform various classification tasks. To mitigate this, various dataset augmentation schemes have been developed \citep{zhao-etal-2018-gender}. Our work uncovers biased features in continuous prompts which can inform when it is appropriate to employ such tactics.

\end{document}